%% file: main.tex
\documentclass[10pt,twocolumn,letterpaper]{article}

\usepackage[pagenumbers]{cvpr} % To force page numbers, e.g. for an arXiv version

\input{preamble}
\definecolor{cvprblue}{rgb}{0.21,0.49,0.74}
\usepackage[pagebackref,breaklinks,colorlinks,allcolors=cvprblue]{hyperref}
\usepackage{pifont}
\usepackage{multirow}
\usepackage[normalem]{ulem}
\useunder{\uline}{\ul}{}
\usepackage{booktabs}
\usepackage{siunitx}
\usepackage{subcaption}
\usepackage{graphicx}
\usepackage[most]{tcolorbox}   % 加载 tcolorbox
\usepackage{listings}          % 可选：代码高亮
\tcbset{
  colback=gray!10,
  colframe=gray!50,
  boxrule=0.3pt,
  arc=2pt,
  left=4pt,
  right=4pt,
  top=4pt,
  bottom=4pt
}
\def\paperID{*****} % *** Enter the Paper ID here
\def\confName{CVPR}
\def\confYear{2026}

\title{LifeEval: A Multimodal Benchmark for Assistive AI in \\ Egocentric Daily Life Tasks}

\author{
Hengjian Gao$^{1,2}$, 
Kaiwei Zhang$^{1}$\thanks{Corresponding author.}, 
Shibo Wang$^{2}$, 
Mingjie Chen$^{4}$, 
Qihang Cao$^{4}$, 
Xianfeng Wang$^{2}$, \\
Yucheng Zhu$^{2}$, 
Xiongkuo Min$^{2}$, 
Wei Sun$^{3}$, 
Dandan Zhu$^{3}$, 
Guangtao Zhai$^{1,2}$\\
$^{1}$Shanghai Artificial Intelligence Laboratory,
$^{2}$Shanghai Jiao Tong University\\
$^{3}$East China Normal University,
$^{4}$Shanghai University of Electric Power\\
{\tt\small \{hengjiang, zhangkaiwei\}@sjtu.edu.cn}
}

\begin{document}
\maketitle
\input{sec/0_abstract}    
\input{sec/1_intro}
\input{sec/2_relatedwork}
\input{sec/3_Dataset}

\input{sec/4_experiment}

\input{sec/5_conclusion}
\input{sec/6_acknowledgement}
{
    \small
    \bibliographystyle{ieeenat_fullname}
    \bibliography{main}
}

%     \small
%     \bibliographystyle{ieeenat_fullname}
%     \bibliography{main}
% }
\end{document}

%% file: sec/0_abstract.tex
\begin{abstract}
The rapid progress of Multimodal Large Language Models (MLLMs) marks a significant step toward artificial general intelligence, offering great potential for augmenting human capabilities. However, their ability to provide effective assistance in dynamic, real-world environments remains largely underexplored. Existing video benchmarks predominantly assess passive understanding through retrospective analysis or isolated perception tasks, failing to capture the interactive and adaptive nature of real-time user assistance.
To bridge this gap, we introduce \textbf{LifeEval}, a multimodal benchmark designed to evaluate real-time, task-oriented human–AI collaboration in daily life from an egocentric perspective. LifeEval emphasizes three key aspects: task-oriented holistic evaluation, egocentric real-time perception from continuous first-person streams, and human–assistant collaborative interaction through natural dialogues. Constructed via a rigorous annotation pipeline, the benchmark comprises 4,075 high-quality question–answer pairs across 6 core capability dimensions. Extensive evaluations of 26 state-of-the-art MLLMs on LifeEval reveal substantial challenges in achieving timely, effective and adaptive interaction, highlighting essential directions for advancing human-centered interactive intelligence.

\end{abstract}

%% file: sec/1_intro.tex
\section{Introduction}

The long-standing ambition of artificial intelligence, achieving Artificial General Intelligence (AGI), has driven a fundamental shift in research priorities \cite{agi}. Rather than focusing on narrowly defined and isolated tasks, recent studies aim to build systems that can understand, learn from, and assist humans in accomplishing complex activities \cite{AIBench,vats2024survey}.  Recent advances in multimodal large language models (MLLMs) \cite{zhang2025large} have opened unprecedented pathways toward this vision. By integrating visual, linguistic, and other modalities within a unified framework, MLLMs have demonstrated strong cross-modal understanding and reasoning capabilities, encompassing tasks such as image captioning \cite{bai2018survey} and visual question answering \cite{wu2017visual}. These developments have laid a solid foundation for more precise, adaptive, and human-centered interaction.

\begin{figure}[!t]
 \centering
 \includegraphics[width=0.45\textwidth, trim=0mm 0cm 0mm 0cm, clip]{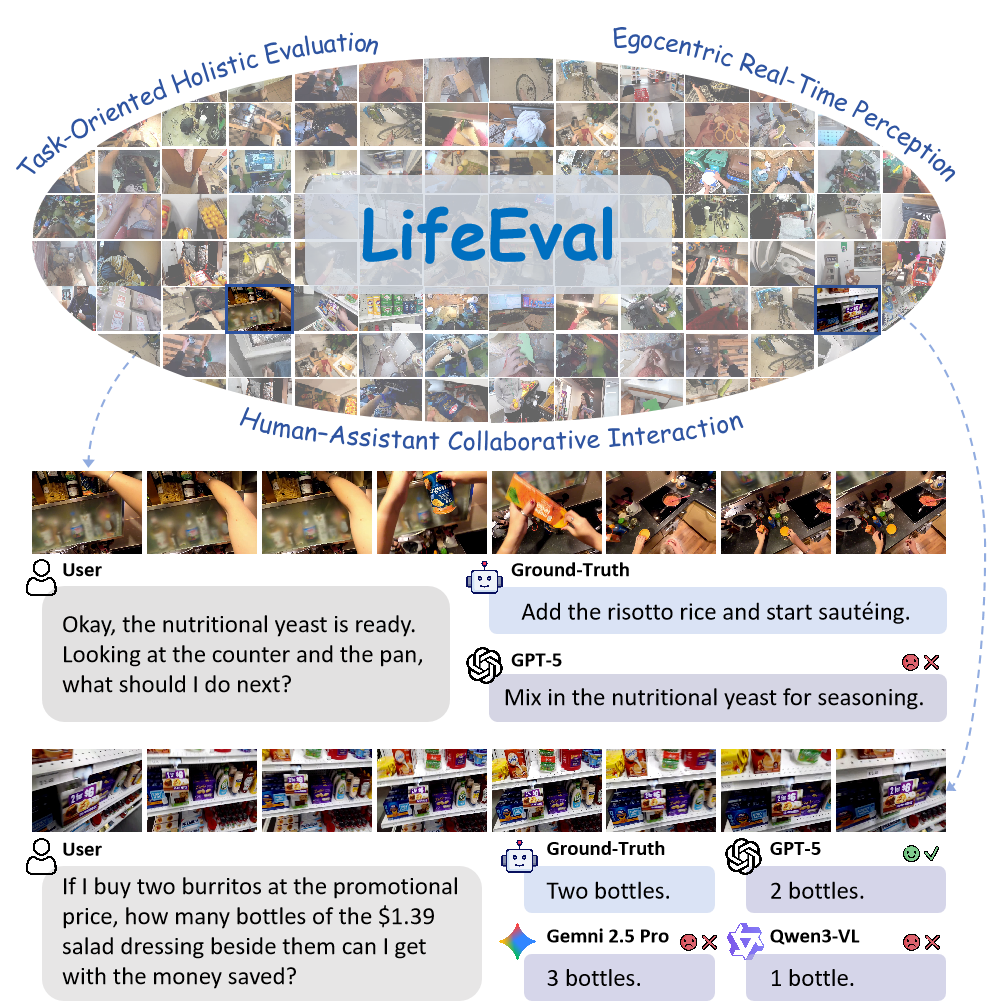}
 % \vspace{-0.1cm}
  \caption{Overview of LifeEval. The benchmark emphasizes real-life interactive scenarios, requiring models to deliver precise, context-aware, and adaptive assistance.}
  % \vspace{-0.5cm}
\label{figure1}
\end{figure}

Meanwhile, as AI technologies become increasingly embedded in daily life through wearable and embodied platforms such as AI glasses and first-person cameras, the focus of intelligence is turning toward user-centered and egocentric interaction \cite{li2025challenges}. In such scenarios, an intelligent assistant must go beyond perceiving and understanding the surrounding environment. It must stay aligned with the user’s perception, intentions, and ongoing activities in real time. This requires the model to analyze egocentric video streams to infer task goals and progress, recognize potential confusion or errors, and provide context-aware and actionable guidance. Unlike traditional third-person video understanding, where the model functions as a detached observer, egocentric interaction requires the model to act as an embedded and proactive participant. This paradigm raises substantially higher requirements for dynamic perception, causal reasoning, and adaptive response, representing a critical step toward the development of truly human-aligned intelligent systems.

To evaluate the video understanding abilities of MLLMs, numerous benchmarks have been developed, covering long-term video memory \cite{omni44,omni59,egoschema}, comprehensive reasoning \cite{mvbench,videott}, and egocentric understanding \cite{videgothink,ecbench}. However, existing benchmarks exhibit two key limitations. First, most benchmarks evaluate models in offline settings using complete video clips, emphasizing retrospective analysis rather than real-time perception and response. Second, their evaluation largely focus on isolated understanding abilities, with limited connection to concrete user goals and intentions. These gaps make it difficult to assess how well models can operate as embedded assistants that must continually align with a user’s evolving intent and provide actionable guidance.
In other words, existing research has focused on "what a model can see and understand", while paying little attention to \textbf{"what a model can actually do to help"}, which is essential for building truly intelligent and useful systems.

To bridge this gap, we introduce \textbf{LifeEval}, a novel benchmark designed to systematically evaluate MLLMs as real-time and task-oriented assistants in first-person scenarios, as illustrated in \cref{figure1}.
The benchmark features three key characteristics: 
(1) \textbf{Task-Oriented Holistic Evaluation.} Instead of assessing isolated perceptual or reasoning skills, we evaluate the model’s holistic ability to assist users throughout task execution, including state tracking, knowledge integration, goal planning, error diagnosis, and actionable guidance.
(2) \textbf{Egocentric Real-Time Perception.} Models are exposed to continuous egocentric video streams, which requires robust online inference to capture semantically meaningful and task-relevant cues under dynamic real-world conditions.
(3) \textbf{Human–Assistant Collaborative Interaction.} By simulating realistic human-assistant dialogues, our benchmark incorporates diverse and context-driven user questions, moving beyond predefined templates to reflect open-ended collaboration.

Through a carefully designed data collection pipeline, LifeEval includes 4,075 high-quality question–answer pairs, covering both multiple-choice and open-ended formats. We further decompose task-assistance capability into 6 key dimensions: static environment perception, dynamic task reasoning, contextual knowledge retrieval, goal-oriented planning, safety and feasibility assessment and multi-turn interactive collaboration. This structure enables a comprehensive and fine-grained evaluation of MLLMs’ performance as real-world assistants.

Based on LifeEval, we conduct extensive benchmarking of a variety of state-of-the-art MLLMs, including both open-source and proprietary models. The results reveal significant gaps in current systems’ ability to provide reliable and efficient assistance in egocentric, real-time settings.
Through LifeEval, we aim to promote research on human-centered multimodal intelligence, and to advance the development of AI systems that can truly understand human intentions and collaborate effectively with people in solving real-world problems.

In summary, our main contributions are:
\begin{itemize}
    \item We propose LifeEval, the first benchmark dedicated to evaluating MLLMs as real-time task assistants from a first-person perspective, marking a shift in focus from passive understanding to active, collaborative interaction.
    \item We design a fine-grained six-dimensional evaluation framework and construct a large-scale, high-quality question-answer bank with detailed rationales via a purpose-built pipeline, ensuring both comprehensiveness and reliability.
    \item We benchmark 26 state-of-the-art MLLMs, revealing their current limitations as intelligent assistants and identifying future directions for advancing human-centered interactive intelligence.
\end{itemize}

\begin{table*}[!t]
\setlength{\abovecaptionskip}{2pt}
% \caption{Comparison of LifeEval with existing video benchmarks. \textbf{\#QAs}: number of question–answer pairs. \textbf{\#Clips}: number of video clips. \textbf{Egocentric}: whether the benchmark is based on egocentric video data. \textbf{Interactive}: whether it simulates human–assistant conversational interaction. \textbf{Real-Time}: whether QA is performed in a real-time or streaming manner. \textbf{Task Assist.}: whether the evaluation focuses on task-oriented assistance. \textbf{Multi-Turn}: presence of multi-round dialogues. \textbf{Rationale}: whether reasoning rationales are provided for each answer. \textbf{Annotation}: the annotation method. Symbols used: \ding{51} (Yes), \ding{109} (Partially), and \ding{55} (No).}
\caption{Comparison of LifeEval with existing video benchmarks. Columns summarize key properties, including dataset size (\#QAs, \#Clips), egocentric perspective, interactive and real-time settings, task-oriented assistance, multi-turn dialogue, availability of reasoning rationales for each answer, and annotation method. Symbols: \ding{51} (Yes), \ding{109} (Partial), \ding{55} (No).}
% 引用
\label{benchcomp}
\centering
\resizebox{\linewidth}{!}{
\begin{tabular}{lcccccccccc}
\toprule
\textbf{Benchmark}      & \textbf{\#QAs} & \textbf{\#Clips} & \textbf{Egocentric} & \textbf{Interactive} & \textbf{Real-Time} & \textbf{Task-Assist.} & \textbf{Multi-Turn} & \textbf{Rationale} & \textbf{Question Type} & \textbf{Annotation}   \\
\midrule
MVBench \cite{mvbench}        & 4000  & 3641    & \ding{55}          & \ding{55}           & \ding{55}         & \ding{55}               & \ding{55}          & \ding{55}         & Close         & Auto         \\
VideoMME \cite{videomme}       & 2700  & 900     & \ding{55}          & \ding{55}           & \ding{55}         & \ding{55}               & \ding{55}          & \ding{55}         & Close         & Manual       \\
Video-TT   \cite{videott}    & 5000  & 1000    & \ding{55}          & \ding{55}           & \ding{55}         & \ding{55}               & \ding{55}          & \ding{51}         & Open\&Close    & Auto\&Manual \\
OmniMMI  \cite{omnimmi}      & 2290  & 1121    & \ding{109}        & \ding{51}           & \ding{51}         & \ding{55}               & \ding{109}        & \ding{55}         & Open          & Manual       \\
OvO-Bench   \cite{ovobench}   & 2814  & 644     & \ding{109}        & \ding{55}           & \ding{109}       & \ding{55}               & \ding{55}          & \ding{55}         & Close         & Auto\&Manual \\
EgoVQA  \cite{egovqa}    & 520  & 520     & \ding{51}          & \ding{55}           & \ding{55}         & \ding{55}               & \ding{55}          & \ding{55}         & Close    & Manual       \\
EgoSchema  \cite{egoschema}    & 5063  & 5063    & \ding{51}          & \ding{55}           & \ding{55}         & \ding{55}               & \ding{55}          & \ding{55}         & Close         & Auto\&Manual \\
EgoPlan-Bench2 \cite{egoplan2} & 1321  & 1113    & \ding{51}          & \ding{51}           & \ding{55}         & \ding{51}               & \ding{55}          & \ding{55}         & Close         & Auto\&Manual \\
EOC-Bench  \cite{eocbench}    & 3277  & 656     & \ding{51}          & \ding{55}           & \ding{55}         & \ding{55}               & \ding{55}          & \ding{55}         & Open\&Close    & Manual       \\
EgoTextVQA   \cite{egotextvqa}  & 7064  & 1507    & \ding{51}          & \ding{51}           & \ding{109}       & \ding{109}             & \ding{55}          & \ding{55}         & Open          & Auto\&Manual \\
EgoLifeQA  \cite{egolife}    & 6000  & 6       & \ding{51}          & \ding{55}           & \ding{55}         & \ding{55}               & \ding{55}          & \ding{51}         & Close         & Manual       \\
VidEgoThink \cite{videgothink}   & 4993  & 3665    & \ding{51}          & \ding{51}           & \ding{55}         & \ding{109}             & \ding{55}          & \ding{55}         & Open          & Auto         \\
\midrule
\textbf{LifeEval} (Ours) & 4075  & 591     & \ding{51}          & \ding{51}           & \ding{51}         & \ding{51}               & \ding{109}       & \ding{51}         & Open\&Close    & Auto\&Manual \\
\bottomrule
\end{tabular}
}
\end{table*}

% 借鉴egotextvqa

%% file: sec/2_relatedwork.tex
\section{Related Works}

% \subsection{Multi-modal Large Language Models}

\subsection{General Video Understanding Benchmarks}

With the rapid advancement of Multimodal Large Language Models (MLLMs) \cite{gpt5,gemini2.5,grok4,qwen3,intern3.5}, evaluating video understanding capabilities has become an increasingly important research focus \cite{zhang2025large}.
% In recent years, a number of representative benchmarks have been introduced to assess different aspects of video comprehension.
Early benchmarks \cite{mvb21,mvb29,mvb81,mvb82} primarily concentrated on specific subtasks such as multimodal retrieval and visual question answering. Subsequent works \cite{omni43,mvb79} extended these efforts to temporal reasoning, and more recent benchmarks \cite{videomme,omni44,omni59} have focused on long-form video understanding. Among the more comprehensive efforts, MMBench-Video \cite{mmbenchvideo} introduced a detailed taxonomy for fine-grained evaluation across multiple dimensions. Video-MME \cite{videomme} incorporated diverse tasks like recognition, perception, and reasoning, to form a holistic evaluation framework. MVBench \cite{mvbench} innovatively formulated spatiotemporal tasks to jointly assess spatial and temporal reasoning within a unified benchmark. 

Despite these advances, most general video understanding benchmarks still lack evaluation of real-time perception and response capabilities, which are essential for interactive understanding. Although several recent benchmarks \cite{omnimmi,svbench} have explored dialogue-based evaluation of streaming video understanding, they are not primarily designed for egocentric scenarios. This mismatch with the way humans naturally perceive and interpret the world limits their applicability to real-world human–AI collaboration.

\subsection{Egocentric Video Understanding Benchmarks}

Compared with third-person videos, egocentric data more closely reflects human visual experience, capturing how individuals actually see and interact with the world.
At the same time, the frequent and dynamic perspective changes introduce unique challenges for visual understanding and reasoning \cite{li2025challenges}. 
The release of large-scale egocentric datasets \cite{ego4d,damen2020epic,egtea,el29} has accelerated progress in this domain, enabling the development of benchmarks that assess diverse aspects of egocentric video understanding.
EgoVQA \cite{egovqa} and EgoTaskQA \cite{egotaskqa} focused on video reasoning and comprehension abilities, while EgoSchema \cite{egoschema} emphasized temporal understanding of modern vision and language systems. EgoMemoria \cite{egomemoria}, HourVideo \cite{hourvideo}, and EgoMem \cite{egomem} extended the temporal scope to study long-range memory modeling and cross-temporal semantic consistency. EgoPlan-Bench \cite{egoplan} and EgoPlan-Bench2\cite{egoplan2} targeted task planning ability. VidEgoThink\cite{videgothink}, EOC-Bench\cite{eocbench} and ECBench\cite{ecbench} introduced systematic multi-dimensional evaluations to quantify embodied cognitive abilities. EgoLifeQA \cite{egolife} further expands the focus to long-term behavioral patterns and complex social interactions.

However, most existing benchmarks primarily emphasize retrospective analysis rather than real-time interactive response, and are weakly coupled to explicit user goals or intentions. In contrast, LifeEval shifts the evaluation paradigm from passive perception to active collaboration, offering a more faithful assessment of dynamic human–assistant interactions, as highlighted in the comparison in \cref{benchcomp}.

%% file: sec/3_Dataset.tex
\section{LifeEval}

\begin{figure*}[!t]
 \centering
 \includegraphics[width=0.95\textwidth, trim=0mm 0cm 0mm 0cm, clip]{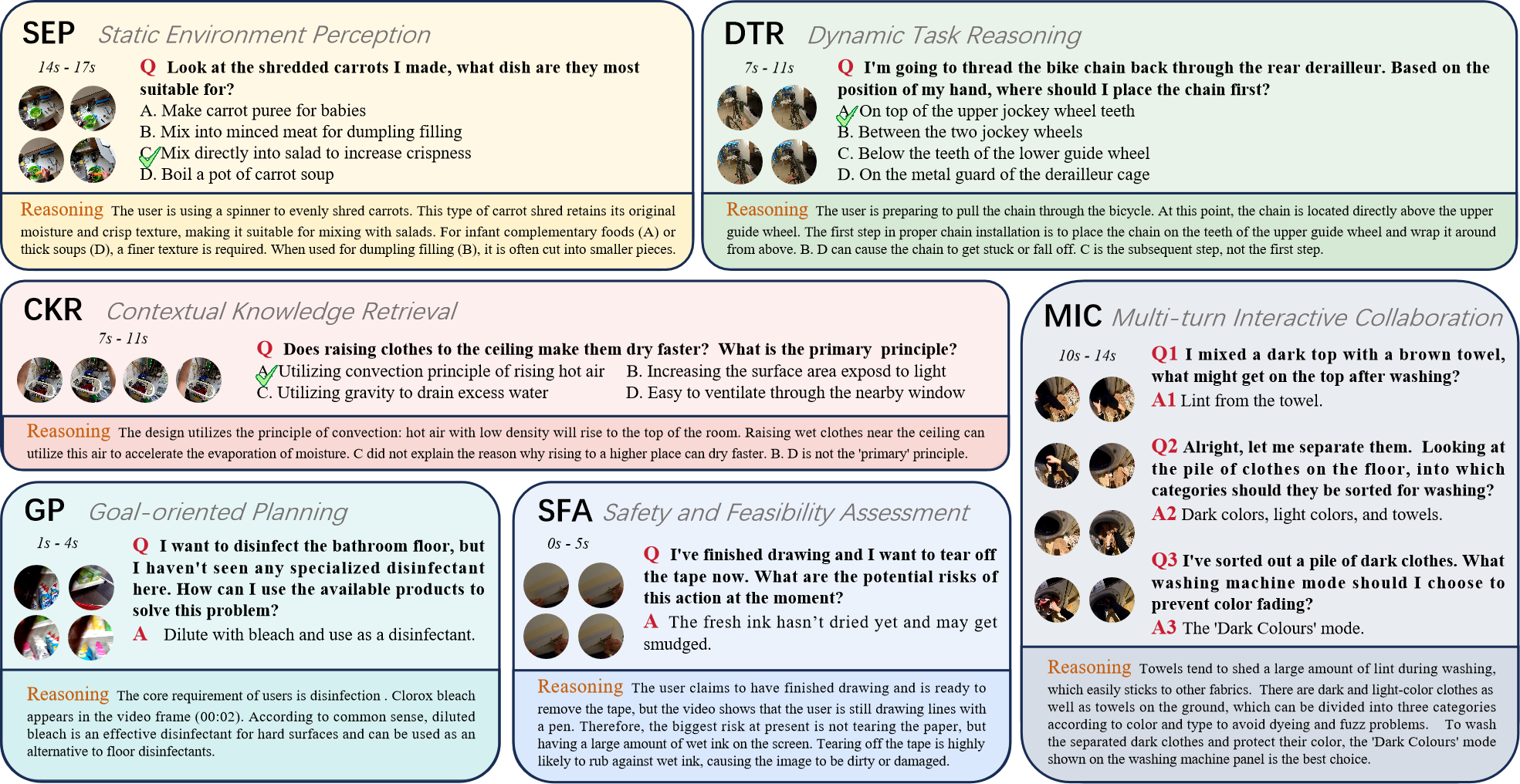}
 % \vspace{-0.1cm}
  \caption{Capability dimensions and examples in LifeEval. LifeEval defines six core capability dimensions to comprehensively evaluate a model’s interactive assistance abilities in daily life. The benchmark features both multiple-choice and open-ended questions, each accompanied by detailed reasoning explanations.}
  % \vspace{-0.5cm}
\label{capaexample}
\end{figure*}

In this section, we first present a hierarchical capability taxonomy for interactive agents. Then, we introduce the construction pipeline and detailed statistics of LifeEval.

\subsection{Capability Taxonomy}
To systematically characterize the competencies required for a multimodal agent to act as an effective collaborative assistant, we introduce a hierarchical taxonomy consisting of six core capabilities. This taxonomy shifts the perspective from passive video understanding to active, context-aware collaboration, emphasizing the correctness, executability, and adaptability of the model’s responses during interaction. It progresses from fundamental perceptual abilities to higher-level interactive reasoning, ensuring a comprehensive evaluation of real-time task assistance, as illustrated in \cref{capaexample}.

\noindent
\textbf{Static Environment Perception (SEP): }
% \paragraph{Static Environment Perception (SEP):}
The ability to recognize and comprehend scene elements that are spatially or instantaneously observable. It focuses on foundational visual grounding, including object identification, attribute recognition, and scene structure comprehension. This dimension captures precise and immediate situational awareness that serves as the perceptual basis for subsequent reasoning and collaboration.

\noindent
\textbf{Dynamic Task Reasoning (DTR): }
% \paragraph{Dynamic Task Reasoning (DTR): }
The capability to infer evolving task progress and state transitions through continuous video streams. It demands temporal reasoning to monitor task evolution, recognize user intentions, infer causal state changes, and predict short-term outcomes. The core of this dimension lies in interpreting the dynamic interplay between the user and the environment over time.

\noindent
\textbf{Contextual Knowledge Retrieval (CKR): }
% \paragraph{Contextual Knowledge Retrieval (CKR): }
The ability to retrieve and integrate external prior knowledge with the current visual context, including commonsense, domain-specific expertise, and procedural manuals. This enables reasoning that goes beyond what is directly visible and allows the model to provide users with additional and instructive guidance.

\noindent
\textbf{Goal-oriented Planning (GP): }
% \paragraph{Goal-oriented Planning (GP): }
The ability to provide effective solutions or next-step guidance based on the current state and the user’s goal. This dimension focuses on the model’s capacity to understand the user’s intention, assess the situation, and deliver actionable instructions that directly assist in problem solving or task advancement.

\noindent
\textbf{Safety and Feasibility Assessment (SFA): }
% \paragraph{Safety and Feasibility Assessment (SFA): }
The ability to evaluate whether the user’s intended actions or current situation are safe and feasible. This includes identifying potential risks, recognizing contraindications, assessing resource availability, and judging the viability of proposed actions, ensuring that the model can act as a reliable safeguard during task execution.

\noindent
\textbf{Multi-turn Interactive Collaboration (MIC): }
% \paragraph{Multi-turn Interactive Collaboration (MIC):}
The integrative ability to maintain coherent and multi-turn dialogue while supporting task completion. This requires preserving contextual consistency, resolving ambiguities, handling coreferences, and collaboratively guiding the user toward task completion. This dimension integrates and evaluates the effective application of all preceding capabilities within a naturalistic conversational flow.

\subsection{Benchmark Construction}

\paragraph{Video Collection}
To evaluate task-oriented assistance, we rely on egocentric video, which closely reflects the user’s natural perspective and interactions with the environment. Ego4D \cite{ego4d} is a representative large-scale egocentric dataset, containing over 3,600 hours of densely annotated videos that spans a wide range of daily activities. Its unprecedented scale and scene diversity make it an ideal source for our benchmark. 

By leveraging both the native annotations and goal-step annotations \cite{song2023ego4d}, we carefully sample a balanced set of videos covering diverse everyday tasks, including but not limited to cooking, cleaning, handicrafts, shopping, laundry and device maintenance. This strategy ensures that our benchmark encompasses the broad spectrum of scenarios an intelligent assistant would encounter in real-world settings. In total, we collect 100 varied daily-life scenarios with a total duration of 44.19 hours.

\paragraph{Data Curation Pipeline}
\label{sec:pipeline}
Following the collection of raw video data, we annotate the videos with high-quality question–answer (QA) pairs in both multiple-choice (MCQ) and open-ended (OEQ) formats, balancing objective evaluation with realistic interactive demands. To achieve cost-effective yet reliable annotation, we adopt a multi-stage pipeline that combines the advanced video reasoning capabilities of Gemini 2.5 Pro \cite{gemini2.5} with targeted human verification, thereby mitigating potential generator-specific priors.

% \noindent
\textbf{(a) MCQ Generation via Direct Video Prompting.} 
Unlike previous works that rely on dense textual narrations to generate QA pairs using text-only LLMs \cite{mvbench, videgothink, egoschema, egoplan, egoplan2}, our approach directly feeds the video content to the video MLLM. This ensures a direct alignment with the raw visual stream, preventing potential information loss or bias introduced by intermediate textual representations. Specifically, we first randomly sample 30-second clips from our collected videos to obtain manageable yet informative segments, followed by manual filtering to remove uninformative ones. For each clip, we employ carefully designed prompts tailored to each of our six capability dimensions, guiding Gemini 2.5 Pro to analyze the scene and generate multiple-choice questions. The model is asked to provide the question, correct answer, a compelling reasoning chain, a set of plausible distractors, and the specific timestamp within the clip that grounds the question, ensuring semantic precision and visual faithfulness.

% \noindent
\textbf{(b) Iterative Quality Filtering.}
To ensure the reliability and precision of the generated questions, we apply a rigorous two-stage quality control process. In the first stage, Gemini 2.5 Pro itself serves as an automated reviewer. Given both the original video clips and the corresponding QA pairs, the model is prompted to identify and revise a variety of issues, including misalignment with the defined capability dimension, unclear phrasing, ambiguous or incorrect answers, weak visual grounding, or redundancy across questions. In the second stage, human annotators conduct a thorough manual review to confirm the clarity, accuracy, and contextual relevance of each QA pair, filtering out any questions that fail to meet our quality standards.

% \noindent
\textbf{(c) Controlled Difficulty Enhancement.}
To further strengthen the benchmark’s discriminative capability, we further introduce a difficulty enhancement stage. In this phase, the questions that pass the previous quality filtering stage are fed back into Gemini 2.5 Pro together with the corresponding video. The model is instructed to reformulate each question to make it more challenging, involving strategies such as requiring multi-step reasoning, incorporating subtle visual cues, or introducing more plausible distractors. The enhanced questions are then reviewed by human annotators to ensure they remain fair, unambiguous, and firmly grounded in the visual content.

% \noindent
\textbf{(d) Open-ended Question Reformulation.}
The final set of high-quality multiple-choice questions serves as the foundation for generating open-ended ones. We prompt Gemini 2.5 Pro to rephrase each MCQ into a concise short-answer question, ensuring that every reformulated question has a clear and semantically unique answer. To preserve answer consistency and evaluability, we exclude questions that are overly broad or subjective. A final round of manual review is then conducted to verify the quality of all reformulated questions.

% \paragraph{Language}

Through this multi-stage pipeline of generation, filtering, enhancement, and reformulation, we finally construct a high-quality benchmark of 4,075 QA pairs, each enriched with reasoning chains and precise temporal grounding.

\begin{figure*}[!t]
  \centering
  \begin{subfigure}[b]{0.25\linewidth}  % 左边图
    \centering
    \includegraphics[width=\linewidth]{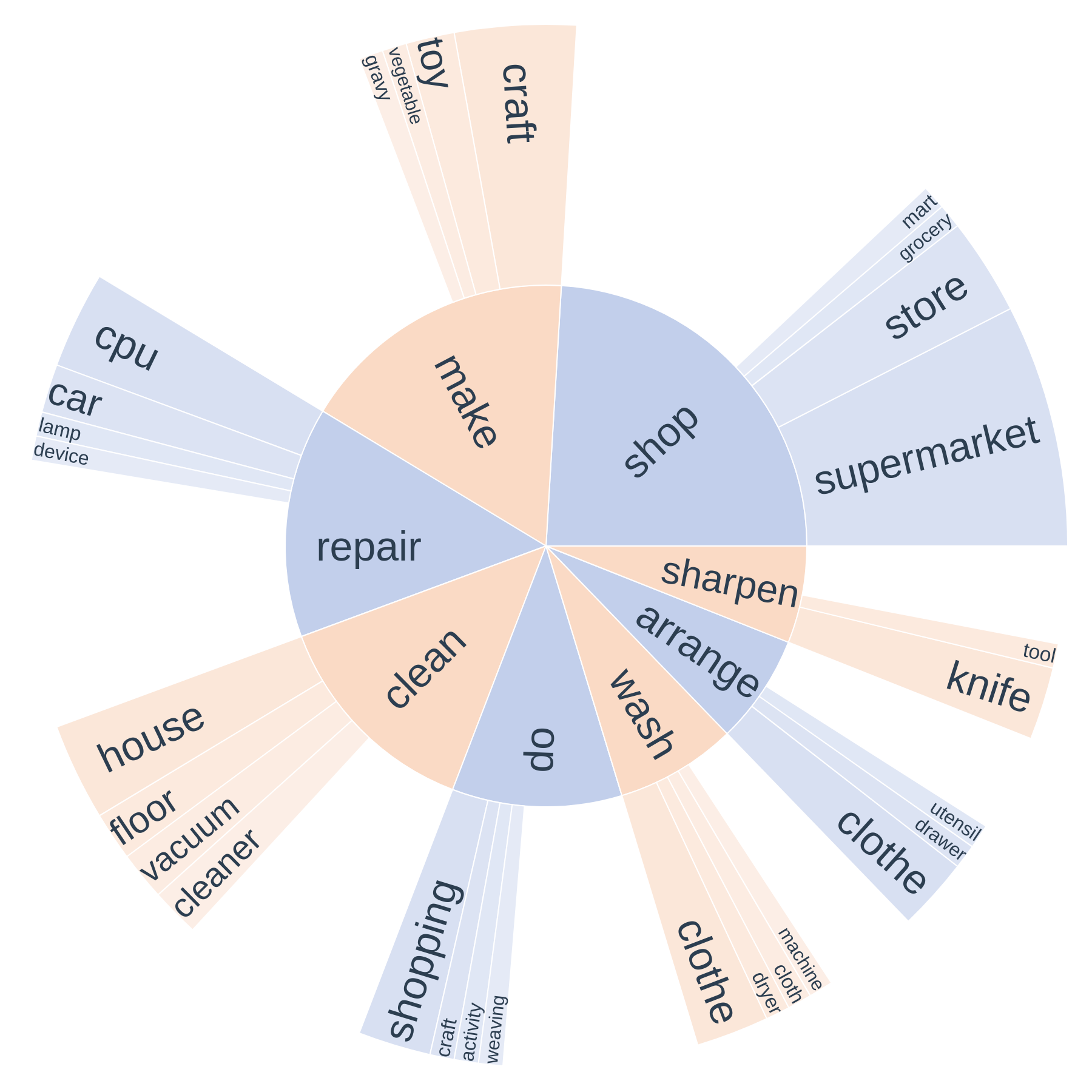}
    % \caption{Top 8 verb–noun co-occurrences in task goals.}
    \caption{Most frequent verb–noun pairs in task goals (Top 8 verbs).}
    \label{fig:sunburst}
  \end{subfigure}\hfill                   % \hfill 把两张图撑开
  \begin{subfigure}[b]{0.35\linewidth}  % 右边图
    \centering
    \includegraphics[width=\linewidth]{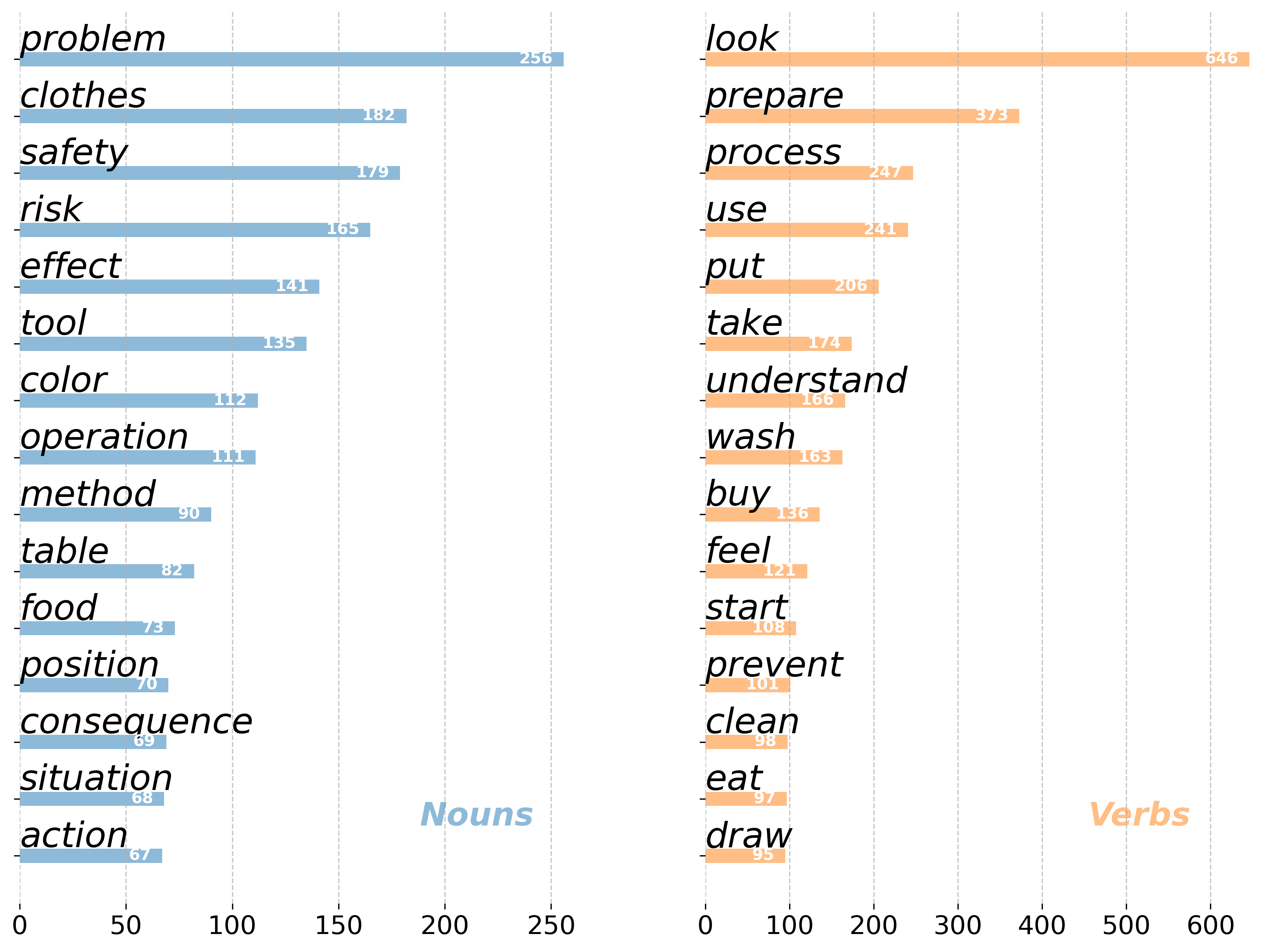}
    \caption{Top 15 most frequent nouns and verbs in questions.}
    \label{fig:question}
  \end{subfigure}\hfill   
  \begin{subfigure}[b]{0.35\linewidth}  % 右边图
    \centering
    \includegraphics[width=\linewidth]{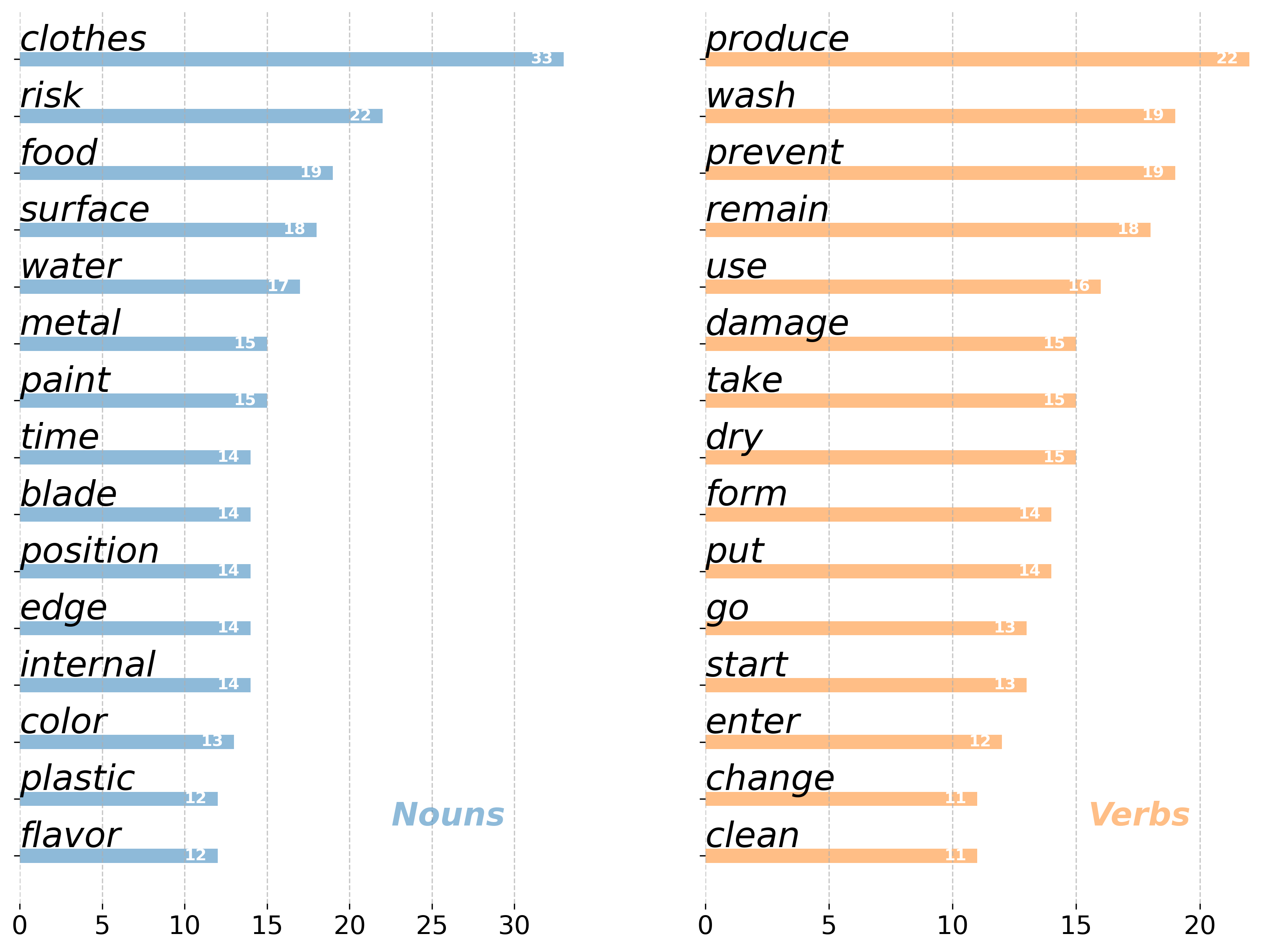}
    \caption{Top 15 most frequent nouns and verbs in answers.}
    \label{fig:answer}
  \end{subfigure}
  \caption{Task goal distribution and QA vocabulary statistics in LifeEval.}
  \label{fig:qagoal}
\end{figure*}

\begin{table}[!t]
\setlength{\abovecaptionskip}{2pt}
\caption{Question statistics of LifeEval.}
\label{questatistic}
\centering
\resizebox{\linewidth}{!}{
\begin{tabular}{lcc}
\toprule
\textbf{Statistics}                                 & \textbf{Number} & \textbf{Percentage} \\
\midrule
Capacities                                &        &            \\
- Static Environment Perception (SEP)        & 666    & 16.34\%    \\
- Dynamic Task Reasoning (DTR)               & 708    & 17.37\%    \\
- Contextual Knowledge Retrieval (CKR)       & 775    & 19.02\%    \\
- Goal-oriented Planning (GP)                & 744    & 18.26\%    \\
- Safety and Feasibility Assessment (SFA)    & 646    & 15.85\%    \\
- Multi-turn Interactive Collaboration (MIC) & 536    & 13.15\%    \\
\midrule
Formats                                  &        &            \\
- Multiple-Choice Questions                & 2069   & 50.77\%    \\
- Open-Ended Questions                     & 2006   & 49.23\%    \\
\midrule
Total                                      & 4075   & 100\%    \\
\bottomrule
\end{tabular}
}
\end{table}

\paragraph{Evaluation Metrics}
\label{sec:evalmetric}
We adopt separate evaluation strategies for multiple-choice and open-ended questions to ensure comprehensive and accurate assessment.
For multiple-choice questions, following standard practice \cite{mvbench, videomme}, we report accuracy by directly matching the model’s response to the ground-truth answer.

For open-ended questions, evaluation focuses on the semantic correctness of free-form text responses.  To achieve this, we use the LLM-as-a-Judge paradigm \cite{gen26, gen3}, where GPT-5 \cite{gpt5} is leveraged as our judge model. In our evaluation framework, the judge compares the model’s answer with the ground-truth reference and assigns a score on a scale from 0 to 1 in 0.25 increments. This graded scoring scheme offers a more nuanced and precise assessment than binary accuracy, effectively capturing partial correctness and subtle differences in answer quality.

\subsection{Benchmark Statistics}
LifeEval comprises 4,075 QA pairs across 591 video clips, spanning two question formats and six capability dimensions to systematically capture everyday human–AI collaboration scenarios. As summarized in \cref{questatistic}, the QA pairs are evenly distributed across both question types and capability dimensions, ensuring a balanced and fine-grained assessment of MLLMs’ interactive abilities. 

To further characterize the benchmark composition, we examine the task goals and linguistic distributions of the QA pairs in \cref{fig:qagoal}. Specifically, we analyze the goal descriptions \cite{song2023ego4d} of all video scenes, as visualized in \cref{fig:sunburst}, which highlights the eight most frequent verbs and their corresponding high-frequency nouns. The results reveal a diverse and balanced coverage of common daily activities, confirming that LifeEval encompasses a broad spectrum of realistic and representative real-world scenarios. 
Moreover, \cref{fig:question,fig:answer} presents the overall word frequency distribution across all QA pairs. We observe that answers tend to contain more specific and concrete nouns, reflecting their goal of providing precise guidance. In contrast, questions exhibit a broader and more diverse vocabulary, including colloquial high-frequency terms such as “look” and “feel”. This pattern indicates that LifeEval effectively captures the natural conversational style of everyday human–assistant interactions rather than artificially constrained or overly formal queries.

%% file: sec/4_experiment.tex
\section{Experiments}

\begin{table*}[!ht]
\setlength{\abovecaptionskip}{2pt}
\caption{Main evaluation results of representative MLLMs on LifeEval across different capability dimensions and question types. \textbf{Bold} and {\ul underline} denote the best and second-best results, respectively.}
\label{tab:main_results}
\centering
\resizebox{\linewidth}{!}{
\begin{tabular}{l|ccccccc|ccccccc}
\toprule
\multirow{2}{*}{\textbf{Model}}     & \multicolumn{7}{c|}{\textbf{Multiple-Choice Questions}}                                                                                              & \multicolumn{7}{c}{\textbf{Open-Ended Questions}}                                                                                            
\\ \cmidrule(lr){2-8} \cmidrule(lr){9-15} 
                           & \textbf{SEP}            & \textbf{DTR}            & \textbf{CKR}            & \textbf{GP}             & \textbf{SFA}            & \textbf{MIC}            & \textbf{Overall}        & \textbf{SEP}            & \textbf{DTR}            & \textbf{CKR}            & \textbf{GP}             & \textbf{SFA}            & \textbf{MIC}            & \textbf{Overall}        \\ \midrule
\multicolumn{15}{c}{\textit{Proprietary MLLMs}}                                                                                                                                                                                                                                   \\ \midrule
GPT-5   \cite{gpt5}                  & {\ul 79.10}    & \textbf{79.73} & {\ul 88.21}    & {\ul 76.66}    & {\ul 84.45}    & {\ul 87.92}    & {\ul 82.48}    & 63.67          & \textbf{66.05} & \textbf{80.65} & \textbf{69.89} & \textbf{79.80} & \textbf{73.92} & \textbf{72.39} \\
GPT-4o \cite{gpt4o}                    & 68.06          & 70.00          & 81.54          & 67.11          & 77.74          & 82.87          & 74.23          & 46.15          & 48.08          & 61.23          & 52.72          & 64.23          & 59.29          & 55.19          \\
GPT-5-mini   \cite{gpt5}              & 77.31          & 75.68          & 84.36          & 75.07          & 82.01          & 86.25          & 79.85          & {\ul 64.95}    & 58.21          & 73.18          & 63.69          & 75.79          & 69.12          & 67.44          \\
Gemini-2.5-Flash \cite{gemini2.5}          & 71.04          & 70.00          & 85.38          & 72.41          & 82.62          & 81.32          & 76.98          & 56.12          & 57.91          & 71.30          & 59.54          & 72.56          & 67.23          & 64.05          \\
Gemini-2.5-Pro   \cite{gemini2.5}          & \textbf{83.28} & {\ul 79.46}    & \textbf{91.54} & \textbf{81.70} & \textbf{89.02} & \textbf{89.68} & \textbf{85.61} & \textbf{65.86} & {\ul 60.58}    & {\ul 74.16}    & {\ul 65.94}    & {\ul 77.75}    & {\ul 72.31}    & {\ul 69.32}    \\
Grok-4    \cite{grok4}                 & 71.94          & 67.57          & 85.38          & 67.90          & 79.88          & 80.27          & 75.30          & 46.22          & 51.33          & 69.68          & 54.70          & 72.01          & 67.62          & 60.07          \\ \midrule
\multicolumn{15}{c}{\textit{Open-Source MLLMs}}                                                                                                                                                                                                                                   \\ \midrule
Qwen3-VL-30B-A3B \cite{qwen3}          & 60.39          & 65.12          & 75.78          & 66.04          & 75.87          & {\ul 78.09}    & 70.01          & \textbf{46.95} & {\ul 40.54}    & {\ul 51.64}    & {\ul 41.77}    & {\ul 60.69}    & \textbf{52.39} & {\ul 48.61}    \\
Qwen3-VL-30B-A3B(thinking) \cite{qwen3} & {\ul 61.49}    & 65.14          & \textbf{76.92} & {\ul 67.37}    & \textbf{78.66} & \textbf{78.78} & {\ul 71.09}    & {\ul 41.41}    & \textbf{44.88} & \textbf{53.31} & \textbf{44.28} & \textbf{62.54} & {\ul 50.41}    & \textbf{49.24} \\
Qwen3-VL-8B \cite{qwen3}               & 52.84          & 54.59          & 64.10          & 55.97          & 66.77          & 70.35          & 60.33          & 37.84          & 36.32          & 42.92          & 37.60          & 47.25          & 42.41          & 40.61          \\
Qwen3-VL-8B(thinking) \cite{qwen3}     & 59.10          & {\ul 65.95}    & 72.05          & 65.25          & 73.48          & 74.44          & 68.16          & 39.50          & 39.20          & 47.34          & 41.69          & 57.00          & 45.78          & 44.96          \\
InternVL3.5-38B  \cite{intern3.5}          & 57.01          & 54.86          & 63.33          & 53.32          & 60.98          & 71.72          & 59.69          & 29.38          & 30.10          & 32.01          & 29.02          & 36.40          & 36.41          & 31.99          \\
InternVL3.5-14B   \cite{intern3.5}         & 55.52          & 57.03          & 61.54          & 50.93          & 53.96          & 67.38          & 57.38          & 26.74          & 25.89          & 28.77          & 22.34          & 36.01          & 34.24          & 28.65          \\
InternVL3.5-8B  \cite{intern3.5}           & 53.13          & 60.00          & 60.00          & 52.79          & 58.84          & 64.56          & 57.98          & 23.87          & 25.74          & 24.29          & 20.64          & 30.90          & 28.44          & 25.40          \\
InternVL3.5-4B  \cite{intern3.5}           & 48.06          & 51.89          & 54.87          & 50.40          & 53.05          & 63.32          & 53.23          & 25.45          & 23.22          & 22.08          & 20.91          & 27.75          & 27.78          & 24.27          \\
LLaVA-OneVision-72B \cite{llavaov}       & \textbf{65.37} & \textbf{70.00} & {\ul 75.90}    & \textbf{72.41} & 72.56          & 77.54          & \textbf{72.19} & 35.05          & 36.24          & 43.96          & 35.76          & 47.25          & 38.49          & 39.48          \\
LLaVA-OneVision-1.5-8B \cite{llavaov1.5}    & 51.64          & 51.14          & 66.00          & 50.99          & 71.80          & 52.97          & 57.47          & 17.74          & 19.69          & 28.22          & 27.63          & 33.33          & 14.42          & 23.83          \\
LLaVA-OneVision-7B \cite{llavaov}        & 56.42          & 53.24          & 61.79          & 53.32          & 51.52          & 61.43          & 56.17          & 29.91          & 25.74          & 31.95          & 29.16          & 32.55          & 29.43          & 29.81          \\
mPLUG-Owl3-7B   \cite{mplug}           & 49.55          & 57.03          & 66.41          & 53.85          & 61.28          & 32.96          & 54.55          & 21.37          & 25.67          & 26.75          & 21.53          & 32.15          & 27.36          & 25.66          \\
mPLUG-Owl3-2B  \cite{mplug}            & 34.93          & 41.35          & 49.23          & 41.64          & 47.87          & 28.56          & 41.22          & 14.73          & 11.09          & 11.88          & 13.42          & 11.48          & 6.91           & 11.78          \\
LLaVA-NeXT-110B \cite{llavanext}           & 55.22          & 60.81          & 70.26          & 59.95          & {\ul 76.22}    & 72.21          & 65.45          & 26.51          & 32.32          & 34.87          & 34.20          & 48.19          & 34.28          & 34.97          \\
LLaVA-NeXT-72B  \cite{llavanext}           & 54.93          & 62.16          & 70.77          & 64.19          & 73.78          & 73.20          & 66.26          & 30.89          & 33.51          & 41.43          & 37.19          & 49.06          & 37.27          & 38.24          \\
LLaVA-NeXT-7B \cite{llavanext}             & 40.90          & 44.32          & 42.31          & 40.85          & 52.13          & 53.31          & 45.16          & 11.63          & 9.32           & 10.00          & 10.56          & 17.22          & 11.55          & 11.61          \\ \midrule
\multicolumn{15}{c}{\textit{Open-Source Video-Specialized LMs}}                                                                                                                                                                                                                   \\ \midrule
VideoLLaMA3-7B  \cite{videollama3}           & 49.25          & 50.54          & {\ul 62.05}    & 50.66          & {\ul 58.54}    & 53.13          & 54.13          & 23.26          & 21.82          & 25.91          & 23.23          & 31.53          & 14.71          & 23.69          \\
LLaVA-Video-7B \cite{llavavideo}            & \textbf{54.93} & {\ul 57.03}    & \textbf{63.33} & \textbf{56.23} & 56.40          & 63.07          & \textbf{58.42} & \textbf{26.44} & \textbf{26.78} & {\ul 28.25}    & {\ul 27.86}    & {\ul 33.96}    & {\ul 28.82}    & {\ul 28.61}    \\
LongVA-7B  \cite{longva}                & 48.96          & 52.43          & 61.03          & {\ul 54.64}    & \textbf{62.80} & \textbf{64.25} & {\ul 57.07}    & 19.11          & 25.30          & 27.79          & 23.98          & \textbf{35.30} & 28.06          & 26.47          \\
InternVideo2.5\_Chat\_8B \cite{internvideo}  & {\ul 53.13}    & \textbf{58.38} & 56.67          & 54.11          & 57.93          & {\ul 63.26}    & 56.99          & {\ul 25.98}    & {\ul 26.55}    & \textbf{29.87} & \textbf{28.07} & 33.73          & \textbf{29.53} & \textbf{28.91} \\
\bottomrule
\end{tabular}
}
\end{table*}

\subsection{Experimental Setup}
To establish comprehensive performance baselines, we evaluate 26 leading MLLMs on LifeEval.
% This selection encompasses a diverse range of models, including both proprietary and open-source models, as well as specialized video understanding models.
Specifically, we evaluate six state-of-the-art proprietary MLLMs: GPT-5 \cite{gpt5}, GPT-4o \cite{gpt4o}, GPT-5-mini \cite{gpt5}, Gemini-2.5-Pro \cite{gemini2.5}, Gemini-2.5-Flash \cite{gemini2.5}, and Grok-4 \cite{grok4}. For open-source MLLMs, five prominent model families are evaluated across multiple parameter scales, including Qwen3-VL \cite{qwen3}, InternVL3.5 \cite{intern3.5}, LLaVA-OneVision \cite{llavaov}, LLaVA-NeXT \cite{llavanext}, and mPLUG-Owl3 \cite{mplug}, each assessed across multiple parameter scales. Furthermore, we also include four video-specialized models: VideoLLaMA3 \cite{videollama3}, LLaVA-Video \cite{llavavideo}, LongVA \cite{longva}, and InternVideo2.5 \cite{internvideo}.

% We adopt a streaming-style evaluation protocol: for each query, the model receives only the visual content within the corresponding temporal window rather than the entire video. All models are evaluated in a zero-shot setting with their default configurations. For models without native video support, we uniformly sample 8 frames from the same short interval. Given our emphasis on real-time responsiveness and the inherently brief intervals, this sampling strategy provides representative yet concise visual input while ensuring fair comparison, as confirmed in subsequent analyses. To further ensure consistent evaluation, we enforce a rule-based output format requiring models to return answers within predefined tags.

We adopt a streaming-style evaluation protocol: for each query, the model receives only the visual content within the corresponding temporal window rather than the full video. All models are evaluated in a zero-shot setting with default configurations. For models without native video support, we uniformly sample 8 frames from the same interval. Given our emphasis on real-time responsiveness and the brief intervals, this strategy provides representative yet concise visual input while ensuring fair comparison, as validated in subsequent analyses. To ensure consistent evaluation, we enforce a rule-based output format requiring answers within predefined tags.

\subsection{Main Results}
In this section, we provide a detailed comparison and analysis. \cref{tab:main_results} reports performance across all capability dimensions and question types, with all scores linearly normalized to 0–100 for consistent comparison.

\paragraph{Proprietary MLLMs.}
Proprietary MLLMs show consistently strong performance, with GPT-5 and Gemini-2.5-Pro achieving the highest overall results. On multiple-choice questions, Gemini-2.5-Pro achieves 85.61\% accuracy, outperforming the best open-source MLLM by 13.42\% and the best video-specialized LM by 27.19\%. For open-ended questions, GPT-5 leads with a score of 72.39\%, exceeding the best open-source and video-specialized models by 23.15\% and 43.48\%, respectively. 
Analysis across capability dimensions reveals a notable performance gap. These models exhibit high accuracy in Contextual Knowledge Retrieval (CKR), demonstrating a robust capacity for accessing and integrating external knowledge. However, their performance declines in Dynamic Task Reasoning (DTR) and Goal-oriented Planning (GP). This contrast underscores a critical weakness in procedural reasoning and adaptive planning, both of which are essential capabilities for intelligent task execution.

\paragraph{Open-Source MLLMs.}
Among open-source MLLMs, the Qwen3-VL series achieves the best overall performance. Although LLaVA-OneVision-72B attains the highest accuracy on multiple-choice questions, its performance on open-ended responses remains less competitive, suggesting limitations in generating precise and semantically grounded answers. The LLaVA-NeXT series also benefits from its large parameter scale, delivering stable yet suboptimal results compared with Qwen3-VL. 
Overall, a substantial performance gap persists between open-source and proprietary models across all evaluated capabilities. This disparity is particularly evident in open-ended question answering, where all open-source models fall short of a 50\% accuracy threshold. These results indicate that current open-source MLLMs still lack the robustness required to deliver precise, context-aware, and actionable guidance necessary for effective task-oriented collaboration.

\paragraph{Open-Source Video-Specialized LMs.}
The video-specialized LMs achieve relatively close performance, with LLaVA-Video obtaining the highest score on multiple-choice questions and InternVideo2.5 slightly outperforming others on open-ended questions. 
When compared to open-source MLLMs of similar parameter size, it becomes evident that the main constraint lies not in architectural specialization, but in the limited model capacity itself. Current parameter scales appear inadequate to support the complex understanding and reasoning required for effective human–AI collaboration in real-world assistive contexts.

% \begin{figure*}[t]
%   \centering
%   \begin{subfigure}[b]{0.6\linewidth}  % 左边图
%     \centering
%     \includegraphics[width=\linewidth]{figures/radar.pdf}
%     \caption{123}
%     \label{fig:left}
%   \end{subfigure}\hfill                   % \hfill 把两张图撑开
%   \begin{subfigure}[b]{0.4\linewidth}  % 右边图
%     \centering
%     \includegraphics[width=\linewidth]{figures/params.png}
%     \caption{123}
%     \label{fig:right}
%   \end{subfigure}
%   \caption{123}
%   \label{fig:side_by_side}
% \end{figure*}

\begin{figure*}[t]
  \centering
  \begin{minipage}[b]{0.6\linewidth}
    \centering
    \includegraphics[width=\linewidth,  trim=0mm 4cm 0mm 0cm, clip]{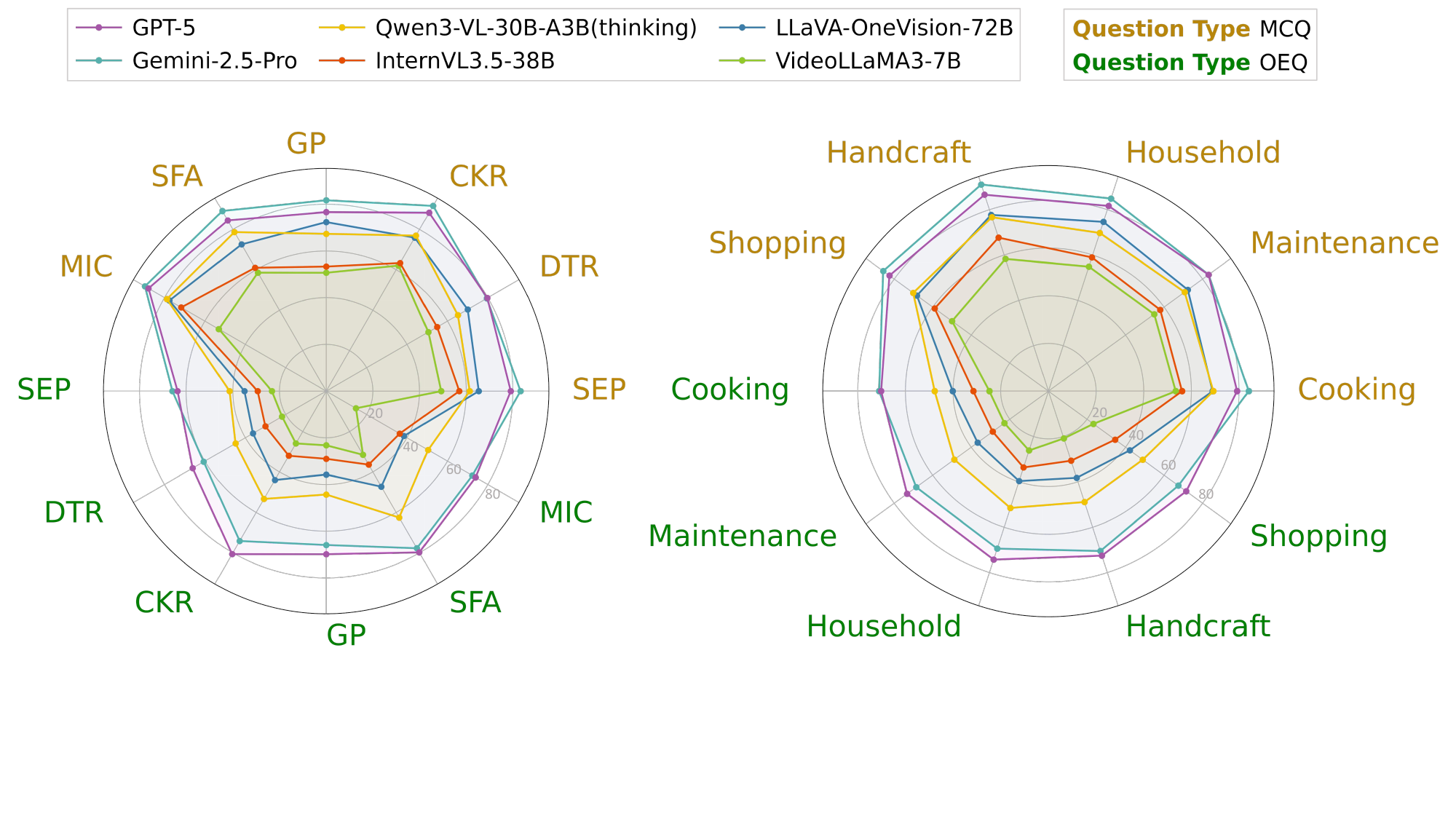}
    \caption{Performance comparison of representative MLLMs on LifeEval. Left: Evaluation results across six capability dimensions. Right: Evaluation results across five everyday scenarios.}
    \label{fig:radar}
  \end{minipage}\hfill
  \begin{minipage}[b]{0.37\linewidth}
    \centering
    \includegraphics[width=\linewidth]{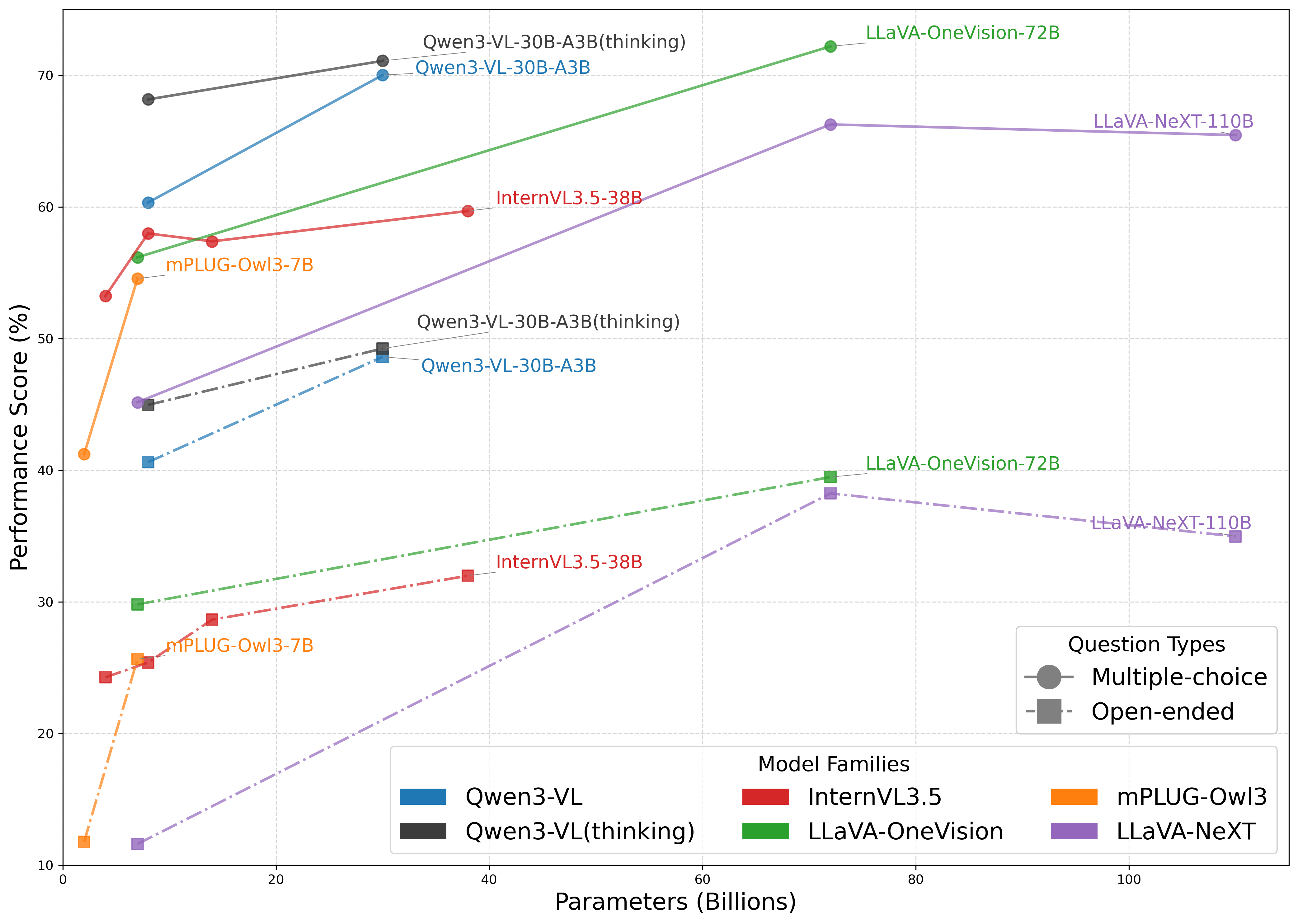}
    \caption{Relationship between model parameter scale and performance on LifeEval for open-source MLLM families.}
    \label{fig:param}
  \end{minipage}
\end{figure*}

\subsection{Further Analysis}
We conduct an in-depth analysis to identify the critical factors influencing model performance as capable task assistants, focusing on capability distribution, scaling effects, and input efficiency.

\paragraph{Performance Gaps Arise from Reasoning and Interaction Challenges Rather Than Scene Variations.}
\cref{fig:radar} illustrates the performance of representative MLLMs across different capability dimensions and everyday scenarios. Overall, GPT-5 and Gemini-2.5-Pro exhibit consistently strong and well-balanced performance across all evaluated aspects, whereas open-source models display more uneven capability distributions. For instance, VideoLLaMA3 shows a notable weakness in Multi-Turn Interactive Collaboration (MIC), while Qwen3-VL performs relatively poorly in Static Environment Perception (SEP).
Besides, all models perform considerably better on multiple-choice questions than on open-ended ones, suggesting that structured options allow models to rely more on recognition, whereas open-ended demand deeper reasoning and generative precision.
Across video scenarios, however, the performance gap remains modest, implying that the primary challenge lies not in scene-specific variations but in the higher-level reasoning and interaction demands of collaborative tasks.

\begin{table}[t]
\setlength{\abovecaptionskip}{2pt}
\centering
\small
\caption{Performance comparison with varying input frame. Numbers in the second row of each model denote performance gain relative to 1-frame.}
\resizebox{\linewidth}{!}{
\begin{tabular}{lcccccc}
\toprule
\multirow{2}{*}{\textbf{\# Frames}} & \multicolumn{3}{c}{\textbf{MCQ}} & \multicolumn{3}{c}{\textbf{OEQ}} \\
\cmidrule(lr){2-4} \cmidrule(lr){5-7}
 & \textbf{1f} & \textbf{8f} & \textbf{32f} & \textbf{1f} & \textbf{8f} & \textbf{32f} \\
\midrule
\multirow{2}{*}{GPT-5} 
 & 80.03 & 82.48 & 82.51 & 67.94 & 72.39 & 70.32 \\
 &  & (+2.45) & (+2.48) &  & (+4.45) & (+2.38) \\
\midrule
\multirow{2}{*}{InternVL3.5-8B} 
 & 56.72 & 57.98 & 57.26 & 24.38 & 25.40 & 26.31 \\
 &  & (+1.26) & (+0.54) &  & (+1.02) & (+1.93) \\
\midrule
\multirow{2}{*}{LLaVA-OV-7B} 
 & 56.65 & 56.17 & 55.39 & 29.11 & 29.81 & 29.22 \\
 &  & ($-$0.48) & ($-$1.26) &  & (+0.70) & (+0.11) \\
\bottomrule
\end{tabular}
}
\label{tab:multiframe_comparison}
\end{table}

\paragraph{Scaling Parameters Alone Is Insufficient for Collaborative Egocentric Tasks.}
We visualize the relationship between model performance and parameter count across various open-source MLLM families in \cref{fig:param}. While model performance generally increases with larger parameter counts across MLLM families, notable exceptions reveal the limitations of pure scaling. For example, in multiple-choice tasks, the InternVL3.5 family exhibits a performance drop when scaling from 8B to 148B parameters, with the 38B variant offering no clear improvement. Similarly, LLaVA-NeXT-110B performs worse than its smaller 72B counterpart. These inconsistencies suggest that larger models may overemphasize complex reasoning or hallucinate irrelevant details, which can be detrimental in egocentric collaboration scenarios. In contrast, Qwen3-VL demonstrates strong performance even with only 30B parameters, highlighting that effective alignment with interaction-oriented and context-grounded understanding tasks can outweigh the benefits of mere model scaling.

\paragraph{Moderate Frame Sampling Suffices for Real-Time Interaction.}
We further investigate the impact of varying input frames numbers on LifeEval, as illustrated in \cref{tab:multiframe_comparison}. Performance improves when increasing from 1 to 8 frames, but further expanding to 32 frames yields no consistent gains and in some cases even degrades results, suggesting that 8 frames represent a near-saturation point for this benchmark. The effect is most pronounced for GPT-5, whereas it is relatively less evident for InternVL3.5-8B and LLaVA-OneVision-7B, likely due to their limited ability to process multiple frames. 
This pattern differs from other benchmarks \cite{eocbench,mmbenchvideo,videomme,longva}, where denser frame sampling often leads to steady improvements. The difference stems from our focus on real-time interaction and cross-domain knowledge integration beyond temporal perception, thereby reducing the benefit of very dense sampling. Excessively frequent frame inputs offer diminishing informational returns while increasing computational latency. These findings highlight a practical trade-off for efficient multimodal assistants: using a moderate number of frames can reduce inference latency while maintaining strong performance, better aligning with real-time interaction demands.

% \paragraph{Error Analysis}
% \paragraph{Analysis of Open-Ended Evalution Metric}

%% file: sec/5_conclusion.tex
% \section{Conclusion}
% In this work, we introduced LifeEval, a comprehensive benchmark for evaluating real-time, task-oriented human–AI collaboration in daily life from an egocentric perspective. Unlike existing video benchmarks that primarily focus on passive perception or offline understanding, LifeEval emphasizes interactive and adaptive assistance in dynamic, real-world contexts. Through 4,075 carefully curated QA pairs spanning 6 fundamental capability dimensions, LifeEval provides a structured yet realistic framework for assessing how well Multimodal Large Language Models perceive, reason, and assist humans in continuous first-person environments.
% Extensive evaluations across 26 state-of-the-art MLLMs reveal that, despite remarkable progress in visual–language integration, several current models still struggle with adaptive reasoning, multi-turn interaction, and contextually grounded collaboration. The performance gap between open-source and proprietary models, as well as the limited benefits from mere scaling or frame sampling, underscores the challenges of translating static understanding into dynamic, real-time assistance.
% By providing a human-centered evaluation framework, LifeEval serves as both a diagnostic tool and a roadmap for advancing multimodal assistants. We hope this benchmark will inspires future research toward more interactive, context-aware, and truly assistive AI systems capable of collaborating with humans seamlessly in everyday life.

\section{Conclusion}
In this work, we introduce LifeEval, a benchmark for evaluating real-time, task-oriented human–AI collaboration from an egocentric perspective. Unlike existing video benchmarks that focus on passive or offline understanding, LifeEval emphasizes interactive assistance in dynamic, real-world scenarios. With 4,075 QA pairs spanning 6 core capability dimensions, it provides a structured framework for assessing how well MLLMs perceive, reason, and assist humans in continuous first-person environments.
Extensive evaluations of 26 state-of-the-art MLLMs show that, despite strong visual–language capabilities, current models still struggle with adaptive reasoning, multi-turn interaction, and grounded collaboration. The gap between models, along with limited gains from scaling or frame sampling, further highlights the challenges of translating static understanding into real-time assistance.
LifeEval offers a human-centered evaluation framework  for advancing more interactive and context-aware multimodal assistants.

%% file: sec/6_acknowledgement.tex
\section{Acknowledgement}
This work was supported by the National Key R\&D Program of China (2025ZD0124104) in collaboration with Shanghai Artificial Intelligence Laboratory, in part by the China Postdoctoral Science Foundation under Grant 2025M781485, and in part by National Natural Science Foundation of China under Grant 62571324.